\documentclass[lettersize,journal]{IEEEtran}
\usepackage{amsmath,amsfonts}
\usepackage{algorithmic}
\usepackage{algorithm}
\usepackage{array}
\usepackage{textcomp}
\usepackage{stfloats}
\usepackage{url}
\usepackage{graphicx}
\usepackage{cite}
\usepackage{amsthm, amssymb}
\usepackage{bm}
\usepackage{bbm}
\usepackage{booktabs}
\usepackage{colortbl}
\usepackage{cellspace}
\usepackage{enumitem}
\usepackage{eqparbox}
\usepackage{footnote}
\usepackage{listings}
\usepackage{multirow}
\usepackage{makecell}
\usepackage{mathtools}
\usepackage{pifont} % \ding
\usepackage{stackengine}
\usepackage{subfigure} 
\usepackage[switch]{lineno}
\usepackage{tabularx}
\usepackage{url}
\usepackage{wrapfig}
\usepackage{xspace}
\usepackage{hyperref} 
\usepackage[dvipsnames]{xcolor}

\usepackage{amsmath,amsfonts,bm}

\def\eqref#1{equation~\ref{#1}}
\def\1{\bm{1}}

\def\vv{{\bm{v}}}

\DeclareMathAlphabet{\mathsfit}{\encodingdefault}{\sfdefault}{m}{sl}
\SetMathAlphabet{\mathsfit}{bold}{\encodingdefault}{\sfdefault}{bx}{n}

\newcommand{\Transpose}{\mathrm{T}}

\theoremstyle{definition}

\theoremstyle{remark}

\newcommand{\xmark}{\ding{55}}%

\newcommand{\av}{{\boldsymbol a}}
\newcommand{\bv}{{\boldsymbol b}}
\newcommand{\cv}{{\boldsymbol c}}

\newcommand{\ev}{{\boldsymbol e}}

\newcommand{\hv}{{\boldsymbol h}}

\newcommand{\lv}{{\boldsymbol l}}

\newcommand{\ov}{{\boldsymbol o}}
\newcommand{\pv}{{\boldsymbol p}}
\newcommand{\qv}{{\boldsymbol q}}

\newcommand{\xv}{{\boldsymbol x}}
\newcommand{\yv}{{\boldsymbol y}}

\newcommand{\thetav}{{\boldsymbol \theta}}
\newcommand{\Amat}{{\bm A}}

\newcommand{\Cmat}{{\bm C}}

\newcommand{\Lmat}{{\bm L}}

\newcommand{\Omat}{{\bm O}}

\newcommand{\Qmat}{{\bm Q}}

\newcommand{\Vmat}{{\bm V}}
\newcommand{\Wmat}{{\bm W}}
\newcommand{\Xmat}{{\bm X}}

\newcommand{\ReLU}{\text{ReLU}}
\newcommand{\BiLSTM}{\text{Bi-LSTM}}

\newcommand{\ModelName}{\text{LiVLR}}
\newcommand{\ModuleName}{\text{DaVL}}

\begin{document}

% title 
\title{LiVLR: A Lightweight Visual-Linguistic Reasoning Framework for Video Question Answering} 

\author{Jingjing~Jiang, 
Ziyi~Liu, 
and~Nanning~Zheng$^*$,~\IEEEmembership{Fellow,~IEEE} 
% % <-this % stops a space 
% \thanks{Manuscript received xxx, 2021; revised xxx, 2021. 
% % This work was supported by the National Natural Science Foundation of China under Grants 61773312, and 62088102. 
% }
\thanks{Jingjing Jiang, Ziyi Liu, and Nanning Zheng are with the Institute of Artificial Intelligence and Robotics, Xi'an Jiaotong University, Shannxi 710049, China (E-mail: jingjingjiang2017@gmail.com, liuziyi@stu.xjtu.edu.cn, nnzheng@mail.xjtu.edu.cn).
}% <-this % stops a space 
\thanks{$^*$Corresponding author. E-mail: nnzheng@mail.xjtu.edu.cn.} 
}

\markboth{Journal of \LaTeX\ Class Files,~Vol.~14, No.~8, August~2021}%
{Shell \MakeLowercase{\textit{et al.}}: A Sample Article Using IEEEtran.cls for IEEE Journals} 

\maketitle

\begin{abstract}
Video Question Answering (VideoQA), aiming to correctly answer the given question based on understanding multi-modal video content, is challenging due to the rich video content. 
From the perspective of video understanding, a good VideoQA framework needs to understand the video content at different semantic levels and flexibly integrate the diverse video content to distill question-related content. 
To this end, we propose a Lightweight Visual-Linguistic Reasoning framework named $\ModelName$. 
Specifically, $\ModelName$ first utilizes the graph-based Visual and Linguistic Encoders to obtain multi-grained visual and linguistic representations. 
Subsequently, the obtained representations are integrated with the devised Diversity-aware Visual-Linguistic Reasoning module (DaVL). 
The DaVL considers the difference between the different types of representations and can flexibly adjust the importance of different types of representations when generating the question-related joint representation, which is an effective and general representation integration method. 
The proposed $\ModelName$ is lightweight and shows its performance advantage on two VideoQA benchmarks, MRSVTT-QA and KnowIT VQA. 
Extensive ablation studies demonstrate the effectiveness of $\ModelName$ key components. 
% Code is available at \url{https://github.com/jingjing12110/LiVLR-VideoQA}. 
\end{abstract}

\begin{IEEEkeywords}
Video question answering, Relational reasoning, Graph convolutional network, Representation integration. 
\end{IEEEkeywords}

%%%%%%%%% BODY TEXT 
\section{Introduction}
\label{sec:introduction}

% PartI : 背景和问题的引入
\IEEEPARstart{V}{ideo} Question Answering (VideoQA) is a typical task of multi-modal understanding, aiming to correctly answer the given question based on understanding video content. 
Due to the rich content, it is challenging to find evidence of the correct answer from the massive video information. 
From the video understanding perspective, a good VideoQA framework wants two crucial functions: 
(\emph{i}) understanding the video content at different semantic levels. 
(\emph{ii}) flexibly integrating the diverse content to distill question-related content.

\begin{figure*}[!t]
\begin{center}
\includegraphics[width=1.0\linewidth]{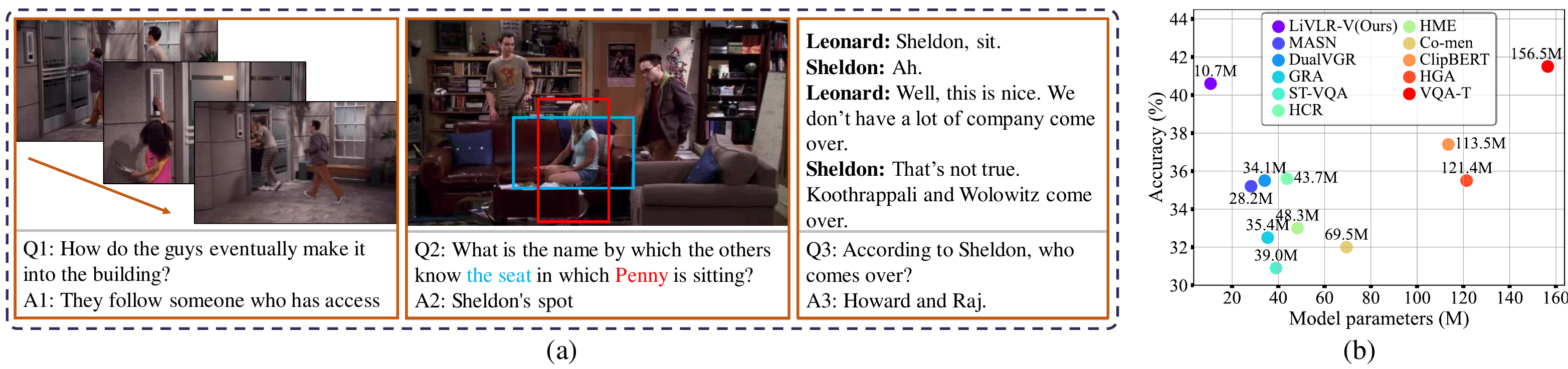}
\end{center}
\vspace{-5mm}
\caption{(a) \textbf{Examples of the VideoQA task}. Answering Q1 requires understanding the holistic event described in the video stream. 
Answering Q2 requires capturing the fine-grained relationship between \textit{the seat} and \textit{Penny} in one frame of the video.  
Answering Q3 requires understanding the linguistic content of the video. 
(b) \textbf{Comparison of the number of model parameters and the accuracy on MSRVTT-QA}. 
The VideoQA models GRA\cite{xu2017video}, ST-VQA\cite{jang2017tgif}, HCR\cite{le2020hierarchical}, HME\cite{fan2019heterogeneous}, Co-mem\cite{gao2018motion}, ClipBERT\cite{lei2021less}, and VQA-T~\cite{yang2021just} only use attention mechanics. 
The models MASA~\cite{seo2021attend}, DualVGR\cite{wang2021dualvgr}, and HGA\cite{jiang2020reasoning} use graph neural networks. 
$\ModelName$-V is a version of our $\ModelName$, which only integrates the multi-grained visual representations in $\ModuleName$. 
}
\label{fig:diversity} 
\end{figure*}

For the first function, the pioneering works~\cite{xu2017video,jang2017tgif,gao2018motion} capture the spatial-temporal information of video and represent them with appearance and motion features. 
These image-level and clip-level representations carry the information needed to answer the types of questions conditioned on holistic video understanding. 
For example, to answer the question Q1 in Figure~\ref{fig:diversity} (a), the VideoQA model requires capturing the holistic event (\ie, the two guys follow the girl into the building) described in the video stream. 
While to answer the type of questions based on video details like Q2 in Figure~\ref{fig:diversity} (a), the VideoQA model needs to identify the seat that Penny is sitting on and to capture the fine-grained relationship between \textit{the seat} and \textit{Penny} in one frame of the video. 
To this end, the relational reasoning-based VideoQA methods~\cite{jin2019multi,huang2020location,seo2021attend} are proposed to model relationships between visual objects. 
In addition to the aforementioned multi-grained visual content, there are some videos contain linguistic content, such as subtitles~\cite{lei2018tvqa,lei2020tvqa}, knowledge~\cite{garcia2020knowledge}, and descriptions~\cite{xu2017video}. 
Analogously, the VideoQA framework also needs to properly understand the holistic and fine-grained linguistic content to answer the questions concerning the linguistic content (like Q3 in Figure~\ref{fig:diversity}~(a)) or even to support visual understanding. 
% In this paper, we consider all cases listed above, and represent the video content to multi-grained visual and linguistic representations. 
Therefore, for a versatile VideoQA framework, it should consider all the cases listed above and flexibly react to each case.

For the second function, \ie, effectively integrating the obtained diverse representations for answer prediction, the existing solutions can be roughly divided into two categories. 
One is attention-based solutions, which design different attention mechanisms, such as memory-enhanced attention~\cite{gao2018motion,fan2019heterogeneous}, spatial-temporal attention~\cite{jang2017tgif,jiang2020divide}, and cross-modality transformer~\cite{lei2021less,yang2021just}, for the fusion of diverse representations. 
The other is attention- and graph-based solutions, which adopt both attention mechanisms and graph reasoning for the representation fusion. 
For example, the works~\cite{jiang2020reasoning,wang2021dualvgr} sequentially apply question-related attention and graph reasoning for the diverse representations fusion. 
However, the attention mechanism includes many matrix multiplication operations with high-dimensional dense representations, which increases the number of model parameters and reduces computational efficiency. 
As shown in Figure~\ref{fig:diversity} (b), the VideoQA model using attention mechanisms to achieve the function (\emph{i}) and (\emph{ii}) like VQA-T~\cite{yang2021just} are usually more heavy-weight than the model that utilizes graph neural networks like DualVGR~\cite{wang2021dualvgr}. 
Therefore, the graph reasoning network is one feasible solution to devise the lightweight VideoQA model.

In this paper, we propose a Lightweight Visual-Linguistic Reasoning framework, named $\ModelName$, which mainly consists of Visual Encoder, Linguistic Encoder, and the Diversity-aware Visual-Linguistic Reasoning module ($\ModuleName$). 
Firstly, $\ModelName$ respectively applies the graph-based Visual Encoder and Linguistic Encoder to encode the visual and linguistic content of the video at different semantic levels and yield multi-grained visual and linguistic representations. 
Subsequently, the obtained multi-grained visual and linguistic representations and the question representation are passed into the Diversity-aware Visual-Linguistic Reasoning module ($\ModuleName$). 
In $\ModuleName$, we construct a diversity-aware graph with the multi-grained visual and linguistic representations as initial node representations. 
The initial node representations are first associated with the question representation using an attention block, and then enhanced by the learnable index embeddings of different representations. 
Facilitated by the learnable embeddings, which prompts the differences of different types of representations and adjusts the importance of different types of representations, $\ModuleName$ can flexibly react to different case of question in using the graph convolutional network to yield a joint representation for answer prediction.

% PartV : 贡献总结 
% \noindent\textbf{Contribution.} 
Our main contributions are summarized as follows: 
\begin{itemize}
\item We propose a Lightweight Visual-Linguistic Reasoning framework for VideoQA, named $\ModelName$, which separately generates multi-grained visual and linguistic representations using graph-based Visual and Linguistic Encoders, and effectively integrates multi-grained visual and linguistic representations via a proposed representation integration method $\ModuleName$. 
\item We propose the Diversity-aware Visual-Linguistic Reasoning module ($\ModuleName$), a powerful and general representation integration method considering the diversity of multi-grained visual and linguistic representations. 
\item The proposed VideoQA framework $\ModelName$ is lightweight and shows its performance advantage on two standard VideoQA benchmarks. 
Extensive ablation studies on key components of $\ModelName$ demonstrate the effectiveness of the proposed framework. 
\end{itemize}

\begin{figure*}[!t]
\begin{center}
\includegraphics[width=1.0\linewidth]{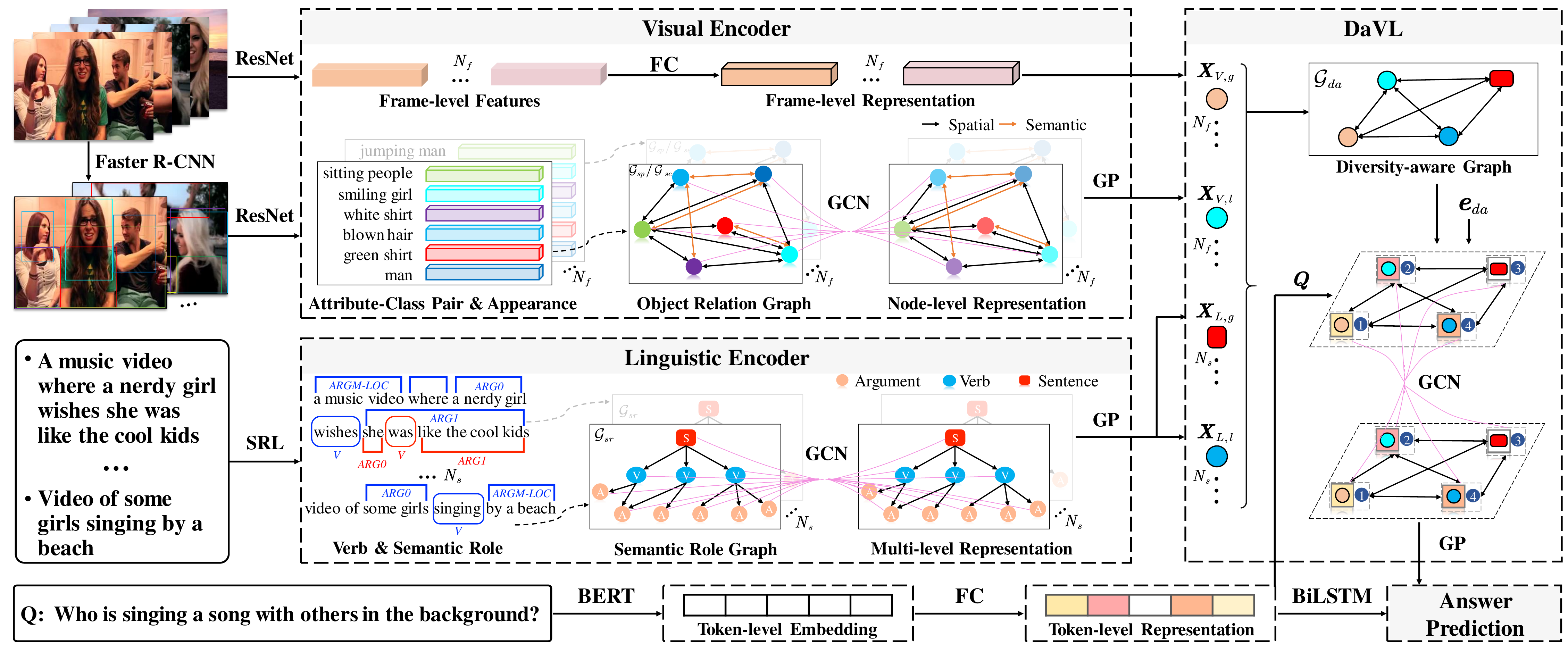}
\end{center}
\vspace{-5mm}
\caption{The overall architecture of $\ModelName$. 
It mainly consists of Visual Encoder, Linguistic Encoder, and the proposed Diversity-aware Visual-Linguistic Reasoning module ($\ModuleName$). 
Visual Encoder separately encodes the holistic and fine-grained visual content and yields multi-grained visual representations. 
Linguistic Encoder uniformly encodes holistic and fine-grained linguistic contents and generates multi-grained linguistic representations.  
$\ModuleName$ aims to integrate multi-grained visual and linguistic representations in a diversity-aware manner and outputs the joint question-related representation for answer prediction. 
}
\label{fig:model}
\end{figure*}

\section{Related Work}
\label{sec:RW}

\subsection{Video Question Answering} 
Video Question Answering aims to answer the given question concerning video content. 
Most current works~\cite{xu2017video,jang2017tgif,gao2018motion,fan2019heterogeneous,yang2019question,li2019learnable,zha2019spatiotemporal} extract holistic visual appearance and motion features to represent video contents and design different attention mechanisms, such as question-guided attention~\cite{xu2017video,jiang2020divide} and co-attention~\cite{gao2018motion,zha2019spatiotemporal}, to integrate these features. 
These methods focus on the holistic understanding of video contents, which may neglect meaningful and fine-grained video contents that complicated semantic questions concern. 

To answer such semantic-complicated questions that are based on fine-grained comprehension of video content, relational reasoning-based methods~\cite{jin2019multi,jiang2020reasoning,le2020hierarchical,huang2020location,kim2020modality,wang2021dualvgr,seo2021attend} are proposed. 
More specifically, Jin \etal~\cite{jin2019multi} propose a multi-modal and multi-level interaction network to capture relations between objects. Jiang \etal~\cite{jiang2020reasoning} develop a heterogenous graph alignment network to integrate the relations of both inter- and intra-modality for cross-modal reasoning. Le \etal~\cite{le2020hierarchical} explore more robust multi-modality interaction by constructing a general-purpose neural reasoning unit. Huang \etal~\cite{huang2020location} propose a location-aware graph convolutional network to model the location and relation among objects explicitly. 
Wang~\etal~\cite{wang2021dualvgr} adopt a stacked dual-visual graph reasoning
unit, DualVGR, to iteratively model rich relationship between video clips. 
Seo~\etal~\cite{seo2021attend} utilize graph convolutional networks to compute the relationships among objects both in appearance and motion modules. 
Park~\etal~\cite{park2021bridge} construct graphs for both video and question and encode question-to-visual relationships and visual-to-visual relationships. 

In addition, to better understand the video content, which is usually a kind of multimodal data including visual and linguistic information, the extra linguistic information, such as subtitles~\cite{lei2018tvqa,lei2020tvqa}, captions~\cite{kim2018multimodal,kim2020dense}, and knowledge~\cite{garcia2020knowit,garcia2020knowledge}, are introduced to VideoQA tasks. 
Our work aims to handle such generalized VideoQA tasks that consider both visual and linguistic information, which is more practical than those visual-specific VideoQA tasks. 
The Visual Encoder in $\ModelName$ is closely related to the relational reasoning-based methods.

\subsection{Relational Reasoning} 
Relational reasoning is extensively exploited in vision-and-language tasks~\cite{yao2018exploring,cadene2019murel,li2019visual,yang2020prior,pei2020visual,chen2020figure} to model intra-modal or cross-modal relations among visual/semantic elements. Recent approaches to relational reasoning can broadly be classified as graph-based~\cite{li2019relation,chen2019graph,gao2020multi}, neuro-symbolic-based~\cite{garcez2019neural,vedantam2019probabilistic,amizadeh2020neuro}, and others~\cite{santoro2017simple,le2020hierarchical,zhang2020multimodal}. 
% Graph-based methods have been proved to be powerful for visual and semantic reasoning and become prevailing methods in vision-and-language tasks, which benefit from the rapid development of Graph Neural Network~\cite{scarselli2008graph}, especially GCN~\cite{kipf2016semi} and GAT~\cite{velivckovic2017graph}. 
Graph-based methods have been proved to be powerful for visual and semantic reasoning and become prevailing methods of relational reasoning in vision-and-language tasks, which usually consider the explicit relation that can be directly denoted by a relation triplet and the implicit relation that is not predefined. 
Specifically, Li \etal~\cite{li2019relation} encode explicit semantic and spatial relations as well as implicit fully-connected relations between objects by a graph-based attention network. Huang \etal~\cite{huang2020location} focus on the location and relations among object interaction and propose a location-aware GCN to model implicit relations between objects. A more recent study~\cite{zhu2020mucko} considers explicit relations in visual, semantic, and knowledge modalities and proposes a modality-aware heterogeneous GCN to encode these relations. 
In addition, the graph learner module~\cite{norcliffe2018learning} conditioned on the context of a given question is developed to better uncover and exploit these implicit relations between objects. 
Similarly, we also employ attention-based GCNs to reveal explicit and implicit relations underlying visual and linguistic content.

\subsection{Graph Neural Network} 
Graph Neural Network (GNN)~\cite{scarselli2008graph} is a type of deep learning models handling graph-structure data, which utilizes the graph structure to aggregate node information from neighborhoods.  
The power of GNNs in modeling relationships between graph nodes makes it widely used in various tasks and applications, such as graph classification~\cite{zhang2018end,bai2020learning}, cross-modality retrieval~\cite{wang2020learning,song2021spatial}, and video question answering~\cite{huang2020location,wang2021dualvgr}. 
In recent years, many GNNs have been proposed. 
These existing GNNs can be broadly grouped into two categories: spectral-based GNNs~\cite{bruna2013spectral,henaff2015deep,rippel2015spectral,defferrard2016convolutional} and spatial-based GNNs~\cite{kipf2016semi,velivckovic2017graph,hamilton2017inductive}. 
Specifically, spectral-based GNNs first transform graphs to the spectral domain by graph Fourier transform, then perform the convolution operator defined in the spectral domain, finally transform the encoded graphs back to spatial domain with the inverse graph Fourier transform. 
For example, Defferrard~\etal~\cite{defferrard2016convolutional} utilize the Chebyshev expansion of the graph Laplacian matrix to define the spectral filters, which alleviates the computational complexity of the eigen-decomposition. 
Spatial-based GNNs directly define convolution operators on the graph based on
the graph topology. 
For example, GAT~\cite{velivckovic2017graph} is a typical spatial-based GNN. 
It incorporates the attention mechanism into the propagation step, which assigns different weights for neighbors to alleviate node noises. 
In this paper, we adopt self-attention based GCNs in Visual and Linguistic Encoders and utilize multi-head attention based GCN in $\ModuleName$.

\section{Preliminary}

% *********************************************************************
In this section, we first state the problem definition and the inputs of $\ModelName$, \ie, the pre-extracted visual and linguistic features. 
After that, we introduce the attention-based GCN, which is the basic block that will be utilized in $\ModelName$. 

\noindent\textbf{Problem Definition:} 
% The goal of the general $\Problem$ task is to infer an answer $\tilde{a}$ for the given question $\mathit{q}$ conditioned on the comprehension of visual content $\mathit{V}$ and linguistic knowledge $\mathit{K}$ about the video. 
The VideoQA task is to infer an answer $\tilde{a}$ for the given question $\mathit{q}$ conditioned on understanding the video content. 
The answer $\tilde{a}$ can be found in an answer set $\mathbb{A}$ that is a predefined set of possible answers for open-ended (OE) question setting or a list of answer candidates for multiple-choice (MC) question setting. 
Since the proposed VideoQA framework $\ModelName$ independently encodes the visual ($\mathit{V}$) and linguistic ($\mathit{L}$) content of the given video, we formulate the VideoQA task as 
\begin{equation} 
\tilde{a} = \underset{a \in \mathbb{A}}{\arg \max}~p_{\thetav}\left(a \mid q,\mathit{V}, \mathit{L}\right), 
\end{equation} 
where $\thetav$ denotes trainable model parameters.

\noindent\textbf{Visual Features:} 
For each video clip, we sample $N_f$ frames of images and represent these sampled images in three forms: 
($i$) image-level appearance features: 
$[\av_{1}^0, \dots, \av_{N_f}^0] \in \mathbb{R}^{N_f \times \text{2048}}$, 
($ii$) object-level region features: 
$[\Omat_{1}^0, \dots, \Omat_{N_f}^0] \in \mathbb{R}^{N_f \times N_o \times \text{2048}}$, 
where, $N_o$ is the number of objects in an image, and $\Omat_f^0$ = $[\ov_1^0, \dots, \ov_{N_o}^0] \in \mathbb{R}^{N_o \times \text{2048}}, 1\leq f\leq N_f$,
and ($iii$) phrase-level class-attribute features: $[\Cmat_{1}^0, \dots, \Cmat_{N_f}^0]\in \mathbb{R}^{N_f \times N_o \times \text{768}}$, where, $\Cmat_f^0$ = $[\cv_1^0, \dots, \cv_{N_o}^0] \in \mathbb{R}^{N_o \times \text{768}}$ is the class-attribute phrase embeddings of the $N_o$ objects in the $f$-th image. 

\noindent\textbf{Linguistic Features:} 
For a given question, we first extract the token-level features $\Qmat^0$ = $[\qv_1^0, \dots, \qv_{N_t}^0] \in \mathbb{R}^{N_t \times \text{768}}$ of the question, where, $N_t$ is the number of tokens in the sentence. 
In addition to the given question, there are $N_s$ linguistic description sentences corresponding to the video-question pair. 
Therefore, we extract linguistic features 
$[\Lmat^0_{1}, \dots, \Lmat^0_{N_s}] \in \mathbb{R}^{N_s\times N_t\times \text{768}}$ for all $N_s$ sentences, where, $\Lmat_s^0$ = $[\lv_1^0, \dots, \lv_{N_t}^0] \in \mathbb{R}^{N_t \times \text{768}}, 1\leq s\leq N_s$. 
That is, the linguistic inputs of $\ModelName$ are the token-level question features and the linguistic description features. 

% *********************************************************************
\noindent\textbf{Attention-based GCN:}
% For brevity, this part introduces attention-based GCN, the basic network utilized in the proposed $\ModelName$. 
For the given graph $\mathcal{G} = (\mathcal{V}, \mathcal{E})$, $\mathcal{V} = \{v_1, ..., v_{N_v}\}$ is a set of nodes, $N_v = |\mathcal{V}|$ denotes the number of nodes, $\mathcal{E}$ is a set of edges, and $\Vmat = [\vv_1, \dots, \vv_{N_v}] \in \mathbb{R}^{N_v\times d}$ is initial node representations. Then, the update formula of node $i$ in the $l$-th GCN layer can be expressed as: 
\begin{equation}
\vv_i^{(l+1)} = \ReLU (\vv_i^{(l)} + \sum_{\vv_j \in \mathcal{N}_i} \alpha_{i,j} \cdot \Wmat^{(l)} \vv_j^{(l)}), 
\label{eq:attention_gcn}
\end{equation}
where $\ReLU$ denotes the ReLU activation function, and $\mathcal{N}_i$ denotes neighborhoods of node $i$, which is determined by $\mathcal{E}$. $\Wmat^{(l)} \in \mathbb{R}^{d \times d}$ is a transformation matrix of node $i$ in $l$-th GCN layer. 
The attention coefficient $\alpha_{i, j}$ is defined as 
\begin{equation} 
\alpha_{i, j} = \frac{\exp((\Wmat_q \vv_i)^{\Transpose} \cdot \Wmat_k \vv_j)}{\sum_{\vv_j \in \mathcal{N}_i}\exp((\Wmat_q \vv_i)^{\Transpose} \cdot \Wmat_k \vv_j)},
\label{eq:attn_factor}
\end{equation}
where, $\Wmat_q \in \mathbb{R}^{d \times d}$ and $\Wmat_k \in \mathbb{R}^{d \times d}$ are learnable transformation matrices.

% ***********************************************************************
% Method 
% ***********************************************************************
\section{Method}
\label{sec:Method}
% 结合框图介绍motivation 
Figure~\ref{fig:model} shows details of the proposed $\ModelName$, consisting of Visual Encoder, Linguistic Encoder, Question Encoder, the proposed $\ModuleName$, and the Answer Prediction module. 
In a nutshell, $\ModelName$ first exploits graph-based encoders to encode fine-grained visual inputs and linguistic inputs and yields multi-grained visual and linguistic representations.  
Subsequently, $\ModelName$ integrates the obtained multi-grained visual and linguistic representations via the Diversity-aware Visual-Linguistic Reasoning module ($\ModuleName$) and yields a joint representation for answer prediction. 
In the following, we sequentially introduce them. 

% ***********************************************************************
\subsection{Visual Encoder}
\label{subsec:ORG}

The Visual Encoder separately encodes the holistic and fine-grained visual contents. 
For each video clip, the image-level appearance features $[\av_1^0, \dots, \av_{N_f}^0]$ are mapped into a $d$-dimensional ($d$-D) space by a fully-connected (FC) layer to obtain the holistic visual representation 
$\Xmat_{V, g}=[\av_1,\dots,\av_{N_f}]\in \mathbb{R}^{N_f \times d}$. 
In order to obtain the fine-grained visual representation, visual relationships between objects are encoded into graphs. 
The specific process is as follows. 

\subsubsection{Object Relation Graph Construction} 
Intuitively, visual relations imply spatial relationships reflecting the relative location of objects and semantic relationships depicting semantic coherence of visual concepts. 
As shown in Figure~\ref{fig:model}, for the $f$-th sampled image, we construct a spatial graph $\mathcal{G}_{sp} = (\mathcal{V}, \mathcal{E}_{sp}, \mathcal{R}_{sp})$ and a semantic graph $\mathcal{G}_{se} = (\mathcal{V}, \mathcal{E}_{se})$ using objects as graph nodes. 
$\mathcal{R}_{sp}$ is a set of edges types. 

\subsubsection{Object Relation Graph Embedding} 
To better represent object-spatial and object-semantic relations, we improve node embeddings of $\mathcal{G}_{sp}$ by concatenating position features and improve node embeddings of $\mathcal{G}_{se}$ by concatenating class-attribute features. 
Specifically, we denote the position feature of object $i$ in the $f$-th image as $\pv_i = [p_x, p_y, p_x + p_w, p_y + p_h, p_w, p_h]^{\Transpose}$, where $(p_x, p_y)$ is the top-left coordinate of the bounding box, $p_w$ and $p_h$ mean the weight and height of the box, respectively. 
Given the feature $\ov_i^0$ of object $i$ in the $f$-th image, the $i$-th node embedding in $\mathcal{G}_{sp}$ can be initialized by 
\begin{align}
% \vv_{sp,i}^{(0)} &= [\Wmat_o\ov_i^0 + \bv_o, \Wmat_{p} \pv_i + \bv_p],
\vv_{sp,i}^{(0)} = \Wmat_{sp}^0([\Wmat_o\ov_i^0 + \bv_o, \Wmat_{p} \pv_i + \bv_p]), 
\label{eq:sp_init}
\end{align}
where, $\Wmat_{o} \in \mathbb{R}^{\text{2048} \times d}$ and $\bv_o \in \mathbb{R}^{d}$ map the extracted object-level feature $\ov_i^0$ into a $d$-D representation. $\Wmat_{p} \in \mathbb{R}^{\text{6} \times d}$ and $\bv_p \in \mathbb{R}^{d}$ map the position feature $\pv_i$ to a $d$-D representation. $\Wmat_{sp}^0 \in \mathbb{R}^{2d\times d}$ transforms the concatenated feature into a $d$-D representation space. 

Given the class-attribute feature $\cv^0_i$ of object $i$ in the $f$-th image, we initialize the $i$-th node embedding in $\mathcal{G}_{se}$ as: 
\begin{align}
\vv_{se,i}^{(0)} = \Wmat_{se}^0([\Wmat_o\ov_i^0 + \bv_o, \Wmat_{c} \cv_i^0 + \bv_c]), 
\end{align}
where, $\Wmat_{o}$ and $\bv_o$ are the same as Eq.~(\ref{eq:sp_init}). 
$\Wmat_{c} \in \mathbb{R}^{\text{768} \times d}$ and $\bv_c \in \mathbb{R}^{d}$ transform $\cv_i^0$ into a $d$-D representation space. $\Wmat_{se}^0 \in \mathbb{R}^{2d\times d}$ transforms the concatenated feature into a $d$-D representation space.

\subsubsection{Object Relation Encoding}
To encode the information of known edge types $\mathcal{R}_{sp}$ into $\mathcal{G}_{sp}$, we modify the information aggregation between node $i$ and its one neighborhood node $j$ in Eq.~(\ref{eq:attention_gcn}) to 
\begin{equation}
\vv_{sp, i\leftarrow j}^{(l)} = \Wmat_{sp}^{(l)} \vv_{sp, j}^{(l)} \oplus \bv_{sp}^{(l)}(r_{i,j}),
\end{equation} 
where, $\Wmat_{sp}^{(l)} \in \mathbb{R}^{d \times d}$ and $\bv_{sp}^{(l)} \in \mathbb{R}^{11}$ are the node transformation matrix and the learnable vector of edge types in the $l$-th GCN layer, respectively. 
$\oplus$ means adding a scalar element-wisely to a vector. 
$r_{i,j} \in \mathcal{R}_{sp}$ indicates the edge type between node $i$ and $j$, which are classified into 11 categories according to recent works~\cite{yao2018exploring,li2019relation}. 
$\bv_{sp}^{(l)}(r_{i,j})$ denotes the $r_{i,j}$-th element of $\bv_{sp}^{(l)}$. 

For $\mathcal{G}_{se}$, considering the complexity of detecting relation triplets between objects in a video, we do not explicitly define the semantic relation but implicitly learn the relation by a graph learner~\cite{norcliffe2018learning}. 
More concretely, the adjacency matrix $\Amat_{se}$ can be obtained using the initial node embeddings $\Vmat_{se}^0$ = $[\vv_{se,1}^{(0)}, \dots, \vv_{se,N_o}^{(0)}] \in \mathbb{R}^{N_o \times d}$: 
\begin{equation}
\Amat_{se} = (\Wmat_1 \Vmat_{se}^0) (\Wmat_2 \Vmat_{se}^0)^{\Transpose},
\label{eq:graph_learner}
\end{equation}
where, $\Wmat_1, \Wmat_2 \in \mathbb{R}^{d \times d}$ are transformation matrices of node embeddings. 
Besides, we adopt a ranking strategy to constrain the graph sparsity, that is, only remain the top $N_n$ maximum values for each row of $\Amat_{se}$. 
After determining the adjacency matrix, the node in $\mathcal{G}_{se}$ can be updated by Eq.~(\ref{eq:attention_gcn}). 

So far, for the $f$-th image in one video clip, we can obtain two node-level representations: $\Vmat_{sp}$ = $[\vv_{sp, 1}, \dots, \vv_{sp, N_o}]\in \mathbb{R}^{N_o\times d}$ from $\mathcal{G}_{sp}$ and $\Vmat_{se}$ = $[\vv_{se, 1}, \dots, \vv_{se, N_o}]\in \mathbb{R}^{N_o\times d}$ from $\mathcal{G}_{se}$. 
After that, we apply graph pooling on the two node-level representations to generate the graph-level embeddings $\bar{\vv}_{sp, f}\in \mathbb{R}^d$ and $\bar{\vv}_{se, f}\in \mathbb{R}^d$, and stack $N_f$ graph-level representations to obtain the fine-grained visual representation $\Xmat_{V, l}$ = $[\xv_1, \dots, \xv_{N_f}]\in \mathbb{R}^{N_f \times d}$, where $\xv_f=\bar{\vv}_{sp,f} + \bar{\vv}_{se,f}$.

% ***********************************************************************
\subsection{Linguistic Encoder} 
\label{subsec:SRG} 
 
For a given video-question pair, there are $N_s$ linguistic sentences. 
We construct a semantic role graph for each sentence. 
The nodes of the semantic role graph include the sentence itself describing an holistic event and the linguistic components in the sentence reflecting fine-grained semantic coherence. 
Therefore, the Linguistic Encoder, which has a similar network architecture to Visual Encoder, uniformly encodes the holistic and fine-grained linguistic contents. 
The specific process is as follows.

\subsubsection{Semantic Role Graph Construction} 
To construct the semantic role graph for the $s$-th ($1 \le s \le N_s$) linguistic sentence, as shown in Figure~\ref{fig:model}, we first adopt an off-the-shelf SRL toolkit~\cite{shi2019simple} to obtain predicates, arguments, and roles of arguments corresponding to the predicates in the sentence. 
With the sentence itself and $N_r$ semantic roles, inspired by works in~\cite{marcheggiani2017encoding,chen2020fine}, we construct the semantic role graph $\mathcal{G}_{sr}=(\mathcal{V}, \mathcal{E}, \mathcal{T}_{sr})$, where, $|\mathcal{V}|=N_r + 1$, and $\mathcal{T}_{sr}$ is a set of node type. 
$\mathcal{G}_{sr}$ is a directed hierarchical graph. 
More specifically, the $s$-th sentence itself serves as a global event node. 
Predicates and arguments are deemed as local action nodes and entity nodes, respectively. 
Each action node is directly connected to the event node, while an entity node is connected with different action nodes according to the semantic role type related to the action node.

\subsubsection{Semantic Role Graph Embedding} 
% 初始化 
For the $\mathcal{G}_{sr}$ of the $s$-th sentence, we initialize the global event node with a sentence-level embedding 
$\lv \in \mathbb{R}^{d}$. 
To obtain the sentence-level embedding, we first use a FC layer to transform the token-level feature $\Lmat_s^0$ of the $s$-th sentence into a $d$-D representation space ($\Lmat_s\in \mathbb{R}^{N_s \times d}$). Then, we apply a one-layer BiLSTM~\cite{hochreiter1997long} on $\Lmat_s$:  
\begin{equation}
\begin{aligned} 
\lv = [\BiLSTM(\overrightarrow{\Lmat_s}; \overrightarrow{\theta_l}); \BiLSTM(\overleftarrow{\Lmat_s}; \overleftarrow{\theta_l})], 
\label{eq:gse_bilstm}
\end{aligned}
\end{equation}
where, $\overrightarrow{\theta_l}$ ($\overleftarrow{\theta_l}$) are the forward (reverse) learned parameters and $[\cdot; \cdot]$ means the concatenation operation. 
For any action/entity node, we initialize it with a token-level feature generated by a non-linear projection: 
\begin{equation} 
\vv_{sr, i}^{(0)} = \Wmat_{sr}^{0} \lv_{t \leftrightarrow i}^0, ~~2 \le i \le N_r + 1, 
\end{equation} 
where, $\lv_{t \leftrightarrow i}^0$ is the token-level feature corresponding to the predicate/argument feature of the $i$-th node in $\Lmat_s^0$. 
$\Wmat_{sr}^{0}\in \mathbb{R}^{\text{768} \times d}$ maps the feature into a $d$-D representation space. 

The semantic role itself implies underlying relationships between the local action node and entity node. 
To introduce the semantic role types into the $\mathcal{G}_{sr}$, we enhance the $i$-th local node in the $l$-th layer with a role embedding, which can be expressed as 
\begin{equation}
\tilde{\vv}_{sr,i}^{(l)} = \vv_{sr,i}^{(l)} \odot \Wmat_{sr}^{(l)}[t_{sr, i}, :], ~2 \le i \le N_r + 1, 
\end{equation}
where, $\odot$ is element-wise multiplication, $\Wmat_{sr}^{(l)} \in \mathbb{R}^{N_r \times d}$ is a learnable role embedding matrix, 
$t_{sr, i} \in \mathcal{T}_{sr} = \{1, ..., N_r \}$ is the semantic role type of node $i$, $\Wmat_{sr}^{(l)}[t_{sr, i}, :]$ denotes the $t_{sr, i}$-th row of $\Wmat_{sr}^{(l)}$.

\subsubsection{Semantic Relation Encoding} 
We employ the attention-based GCN to encode the contextual semantic correlations in $\mathcal{G}_{sr}$. 
Specifically, we first adopt the attention mechanism described as Eq.~(\ref{eq:attn_factor}) to characterize semantic relations of different hierarchical nodes. 
Subsequently, the $i$-th node is updated by the update formula in Eq.~(\ref{eq:attention_gcn}). 

After encoding $\mathcal{G}_{sr}$, we can obtain the event node representation $\vv_{sr, 1}$, which is the generated holistic linguistic representation for the $s$-th sentence. 
We stack $N_s$ event node representations to get the holistic linguistic representations $\Xmat_{L, g}\in \mathbb{R}^{N_s\times d}$. 
To obtain fine-grained linguistic representations of $N_s$ sentences, we first apply an average graph pooling on action and entity node embeddings to gain fined-grained linguistic representation $\bar{\vv}_{sr, s}\in \mathbb{R}^d$ for the $s$-th sentence, and then stack $N_s$ such pooled representations as the fine-grained linguistic representations $\Xmat_{L, l} \in \mathbb{R}^{N_s \times d}$.

% *******************************************
\begin{table*}[!t]
\begin{center}
\caption{Statistics of experimental datasets. 
LType and QType indicate the type of linguistic content provided by the corresponding benchmark and the type of the question setting. 
}
\label{tab:data_statistics} 
\vspace{-2mm}
\small
\setlength{\tabcolsep}{1.2mm}{
% \resizebox{1.0\linewidth}{!}{
% \begin{tabular}{l|rrr|rrr|rrr|ccc|c|c|c}
\begin{tabular}{l|ccc|ccc|ccc|ccc|c|c|c}
% \toprule[1pt]
\hline
\multirow{2}{*}{Dataset} 
&\multicolumn{3}{c|}{\#Question} 
&\multicolumn{3}{c|}{\#Video Clip}  
&\multicolumn{3}{c|}{\#Sentence} 
&\multirow{2}{*}{$N_f$} 
&\multirow{2}{*}{$N_o$} 
&\multirow{2}{*}{$N_s$} 
&\multirow{2}{*}{LType} 
&\multirow{2}{*}{QType} 
&\multirow{2}{*}{Year} \\
% \cmidrule(lr){2-4} \cmidrule(lr){5-7} \cmidrule(lr){8-10} 
\cline{2-10}
~&\multicolumn{1}{c}{Train} 
&\multicolumn{1}{c}{Val} 
&\multicolumn{1}{c|}{Test} 
&\multicolumn{1}{c}{Train} 
&\multicolumn{1}{c}{Val} 
&\multicolumn{1}{c|}{Test} 
&\multicolumn{1}{c}{Train} 
&\multicolumn{1}{c}{Val} 
&\multicolumn{1}{c|}{Test} 
& & & & & \\
% \midrule
\hline
\hline
MSRVTT-QA~\cite{xu2017video} &158,581 &12,278 &72,821 &6,513 &497 &2,990 &78,156 &5,964 &35,880 &64 &10 &12 &caption &OE &2016 \\
KnowIT-VQA~\cite{garcia2020knowit} &19,569 &2,352 &2,361 &9,731 &1,178 &1,178 &19,569 &2,352 &2,361 &32 &12 &12/1 &sub/know &MC &2020 \\
% TVQA~\cite{lei2018tvqa} &122,039 &15,253 &7,623 &17,435 &2,179 &1,089 &17,435 &2,179 &1,089 &32 &12 &16 &Subtitle &MC &2018 \\ 
% \bottomrule[1pt]
\hline
\end{tabular}
}
\end{center}
% \vspace{-2mm}
\end{table*}
% 数据集数字靠右对齐：方便比较大小

% *******************************************

% ***********************************************************************
\subsection{Diversity-aware Visual-Linguistic Reasoning Module}
\label{subsec:fusion}

To better fuse the multi-grained visual and linguistic representations for answer prediction, we consider the diversity of the representations and aligning visual and linguistic representations at different semantic levels (\eg, sentence $\leftrightarrow$ image, semantic roles $\leftrightarrow$ object instances). 
Therefore, we propose to construct a heterogeneous graph with diversity-aware nodes, and utilize a GCN module to further encode and capture relationships between them. 

% Then, the question-related multi-grained visual and linguistic representations are passed into the Diversity-aware Visual-Linguistic Reasoning module ($\ModuleName$) to yield the joint representation. 

\subsubsection{Diversity-aware Graph Construction} 
To integrate the obtained multi-grained visual and linguistic representations in Sec.~\ref{subsec:ORG} and Sec.~\ref{subsec:SRG} in a diversity-aware manner, we construct an undirected heterogeneous graph $\mathcal{G}_{da}$. 
The graph $\mathcal{G}_{da}$ consists of four types of nodes: $N_f$ image-level nodes, $N_f$ object-level nodes, $N_s$ sentence-level nodes, and $N_s$ semantic role-level nodes. 

% we stack them to obtain an ensemble feature $\Vmat^0 \in \mathbb{R}^{N_m \times d}$, where $N_m = 2N_f + 2N_s$, and construct a heterogeneous graph $\mathcal{G}_{gs}$ with $N_m$ nodes initialized by $\Vmat^0$. 

\subsubsection{Diversity-aware Graph Embedding} 
Since the obtained representations $\{\Xmat_{V, g}, \Xmat_{V, l}, \Xmat_{L, g}, \Xmat_{L, l}\}$ are high-level semantic but question-agnostic, we first use an attention block to associate the video content with the given question and to distill question-related representations. 
Specifically, we respectively apply an multi-head attention block~\cite{vaswani2017attention} on the four representations, which can be expressed as: 
\begin{equation}
\begin{aligned}
&\mathcal{Q}_{\text{att}}(\Xmat, \Qmat) = \bigcup_{h=1}^{N_h} \Wmat_h \sigma (\frac{\Wmat_h^q\Xmat(\Wmat_h^k \Qmat)^{\Transpose}}{\sqrt{d_k/N_h}}) \Wmat_h^v \Qmat, 
\label{eq:alpha_ij}
\end{aligned}
\end{equation} 
where, $\Xmat \in \{\Xmat_{V, g}, \Xmat_{V, l}, \Xmat_{L, g}, \Xmat_{L, l}\}$, $\Qmat \in \mathbb{R}^{N_t \times d}$ is the token-level question embedding generated by a non-linear projection that maps $\Qmat^0$ into a $d$-D representation space, 
$\cup$ denotes the concatenation operation in the Eq.~(\ref{eq:alpha_ij}), $N_h$ is the number of heads, $\sigma$ indicates the softmax operation, and $d_k$ is the scaling factor. $\Wmat_q^h, \Wmat_k^h, \Wmat_v^h \in \mathbb{R}^{d/N_h \times d}$ and $\Wmat_x \in \mathbb{R}^{d \times d/N_h}$ are learned parameters. 

The yielded question-related representations are the initial node representations of $\mathcal{G}_{da}$, \ie, $\Vmat^0 \in \mathbb{R}^{(2N_f + 2N_s) \times d}$. 
In addition, to inject the diversity-aware information of multi-grained visual and linguistic representations in representation integration, we use the index embeddings of different types of representations to enhance the initial node representations. 
The index embeddings is learnable and can dynamically adjust the importance of different types of nodes. 
For the $i$-th node in the $0$-th layer, this improvement process can be expressed as: 
\begin{equation} 
\tilde{\vv}_{da, i}^{(0)} = \vv_{da, i}^{(0)} \odot \Wmat_{da}^{(0)}[g_i, :],
\end{equation}
where, $\Wmat_{da}^{(0)} \in \mathbb{R}^{4 \times d}$ is a learnable transformation matrix of the index embedding, $g_i \in \{1, ..., 4\}$ denotes the index of $\{\Xmat_{V, g}, \Xmat_{V, l}, \Xmat_{L, g}, \Xmat_{L, l}\}$, and $\Wmat_{da}[g_i, :]$ denotes the $g_i$-th row of $\Wmat_{da}^{(0)}$. 

\subsubsection{Diversity-aware Graph Encoding} 
There are shallow correlations among the multi-grained and multi-source nodes in $\mathcal{G}_{da}$, such as the temporal correlations among visual nodes and the semantic consistency between visual nodes and linguistic nodes. To delineate these correlations, we apply a vanilla GCN to update the representation of node $i$ in $\mathcal{G}_{da}$: 
\begin{equation}
\vv_{da, i}^{(l+1)} = \ReLU (\tilde{\vv}_{da, i}^{(l)} + \sum_{\tilde{\vv}_{da, j} \in \mathcal{N}_i} \Wmat_{da}^{(l)} \tilde{\vv}_{da, j}^{(l)}),
\label{eq:da_GCN}
\end{equation}
where, $\mathcal{N}_i$ is the neighborhood of node $i$, which is defined by a sparse adjacency matrix learned by Eq.~(\ref{eq:graph_learner}), $\Wmat_{da}^{(l)}\in \mathbb{R}^{d\times d}$ are transformation matrix of node embeddings. 

After effectively encoding these multi-grained and multi-source representations by the granularity- and source-sensitive graph reasoning network, we take an average graph pooling on node embeddings in $\mathcal{G}_{da}$ to obtain the join representation $\hat{\xv} = \bar{\vv}_{da} \in \mathbb{R}^d$ for answer prediction.

% *********************************************************************
\subsection{Question Encoder} 
To further encode the contextual content of the pre-extracted token-level question embedding, we apply a one-layer BiLSTM~\cite{hochreiter1997long} on the token-level question embedding $\Qmat$ to gain the final sentence-level question representation $\hat{\qv} \in \mathbb{R}^d$: 
\begin{equation}
\begin{aligned}
% \overrightarrow{\hv_q} &= \BiLSTM(\overrightarrow{\Qmat^0}; \overrightarrow{\theta_q}), \\
% \overleftarrow{\hv_q} &= \BiLSTM(\overleftarrow{\Qmat^0}; \overleftarrow{\theta_q}), \\ 
\hat{\qv} &= [\BiLSTM(\overrightarrow{\Qmat}; \overrightarrow{\theta_q}); \BiLSTM(\overleftarrow{\Qmat}; \overleftarrow{\theta_q})], 
\label{eq:q_bilstm}
\end{aligned}
\end{equation}
where, $\overrightarrow{\hv_q}$ and $\overleftarrow{\hv_q}$ are the forward and reverse hidden states respectively, $\overrightarrow{\theta_q}$ and $\overleftarrow{\theta_q}$ are learned parameters, and $[\cdot; \cdot]$ means the concatenation operation.

\subsection{Answer Prediction}
\label{subsec:Answer}

\subsubsection{Open-ended} The open-ended question setting is to choose one correct answer from a pre-defined answer set $\mathbb{A}$, which can be regarded as a multi-label classification problem and be trained with a cross-entropy loss function. 
Therefore, we feed the final joint representation $\hat{\xv}$ and the final question representation $\hat{\qv}$ into a classifier with two FC layers ($\mathcal{M}_{\text{cls}}$) to compute label probabilities: 
\begin{equation}
\begin{aligned}
\yv_o = \mathcal{M}_{\text{cls}}([\hat{\xv}; \hat{\qv}]), \yv_o \in \mathbb{R}^{|\mathbb{A}|}.
\end{aligned}
\end{equation}

\subsubsection{Multiple-choice} The multiple-choice question setting is to choose one correct answer from $N_k$ candidates. 
In this case, we first generate the answer embedding $\hat{\ev}_k$ of the $k$-th candidate using a one-layer BiLSTM like Eq.~(\ref{eq:q_bilstm}). 
After that, $\hat{\xv}$, $\hat{\qv}$ and $\hat{\ev_k}$ are fed into a classifier with a linear regression ($\mathcal{M}_{\text{reg}}$) to output the $k$-th answer score: 
\begin{equation}
\begin{aligned}
s_k = \mathcal{M}_{\text{reg}}([\hat{\xv}; \hat{\qv}; \hat{\ev_{k}}]), 1 \leq k \leq N_k,  
\end{aligned}
\end{equation} 
where, the score of the correct candidate is the positive score $s^p$, and the rest scores are negative scores $(s_1^n, \dots, s_{N_k -1}^n)$. 
During training, we utilize the summed pairwise hinge loss $\sum_{t=1}^{N_k-1}\max (0, 1 - (s^p - s_{t}^n))$ between the positive score and each negative score to train our model.

% 背景 - 目的 - 实验设置 - 结果 - 结果分析
\section{Experiments}
\label{sec:Experiments}

% ***********************************************************************
\begin{table*}[!htbp]
\begin{center}
\caption{Comparisons with state-of-the-art methods on MSRVTT-QA. Video representation extractors: ResNeXt-101\cite{hara2018can}, S3D\cite{xie2018rethinking}, C3D~\cite{tran2015learning}, I3D~\cite{carreira2017quo}, BN-Inception\cite{ioffe2015batch}, ResNet50/101/152~\cite{he2016deep}, VGG16\cite{simonyan2014very}, Faster R-CNN~\cite{ren2015faster}, Mask R-CNN\cite{he2017mask}. 
\#Param indicates the number of trainable model parameters. PT means pretraining. 
$\ModelName$ (All) uses extra linguistic inputs (\ie, captions). 
}
\label{tab:MSRVTT} 
\vspace{-2mm}
\small 
\setlength{\tabcolsep}{1.8mm}{
\begin{tabular}{c|l|ccc|c|c|ccccccc}
\hline
\multirow{2}{*}{\#}
&\multirow{2}{*}{Method} 
&\multicolumn{3}{c|}{Video Representation} 
&\multicolumn{1}{c|}{\multirow{2}{*}{\#Param}}
&\multicolumn{1}{c|}{\multirow{2}{*}{PT}}
&\multicolumn{6}{c}{Accuracy (\%)} 
\\ 
& &clip-level &image-level &object-level & &
&What &Who &How &When &Where &All \\ 
\hline
\hline
\multirow{4}{*}{\ding{172}}
& SSML\cite{amrani2021noise} &ResNeXt-101 &ResNet-152 &\xmark &- &\checkmark
&- &- &- &- &- &35.1 \\ 
& ClipBERT\cite{lei2021less} &\xmark &ResNet-50 &\xmark &113.5M &\checkmark 
&- &- &- &- &- &37.4 \\ 
& CoMVT\cite{seo2021look} &S3D &\xmark &Faster R-CNN &- &\checkmark 
&- &- &- &- &- &39.5 \\ 
& VQA-T\cite{yang2021just} &S3D &\xmark &\xmark &156.5M &\checkmark 
&- &- &- &- &- &41.5 \\ 
\hline 
\multirow{6}{*}{\ding{173}} 
& ST-VQA\cite{jang2017tgif} &C3D &ResNet-152 &\xmark &39.0M &\xmark 
&24.5 &41.2 &78.0 &76.5 &34.9 &30.9 \\
% ST-VQA在MSRVTT数据集上的实验结果来源于[HME]
& Co-mem\cite{gao2018motion} &BN-Inception &ResNet-152 &\xmark &69.5M &\xmark 
&23.9 &42.5 &74.1 &69.0 &42.9 &32.0 \\
& GRA\cite{xu2017video} &C3D &VGG16 &\xmark &35.4M &\xmark 
&26.2 &43.0 &80.2 &72.5 &30.0 &32.5 \\
% 通过AMU单元实现 
& HME\cite{fan2019heterogeneous} &C3D &VGG16 &\xmark &48.3M &\xmark 
&26.5 &43.6 &82.4 &76.0 &28.6 &33.0 \\
& MiNOR\cite{jin2019multi} &\xmark &VGG16 &Mask R-CNN &- &\xmark 
&29.5 &45.0 &83.2 &74.7 &42.4 &35.4 \\ 
& HCR\cite{le2020hierarchical} &ResNeXt-101 &ResNet-101 &\xmark &43.7M &\xmark 
&- &- &- &- &- &35.6 \\ 
% HME 代码复现了之前的ST和Co-mem 
% STA\cite{gao2019structured} &- &ResNet-152 &- &- &\xmark 
% &27.9 &43.1 &82.7 & 73.7 & 37.2 & 33.8 \\ 
% % STA在MSRVTT数据集上的实验结果由论文[MiNOR]提供 
% QueST\cite{jiang2020divide} &- &ResNet-152 &- &- &\xmark 
% &27.9 &45.6 &83.0 &75.7 &31.6 &34.6 \\
% % QueST 在四个数据集上进行了实验 
\hline
\multirow{6}{*}{\ding{174}}
& MASN\cite{seo2021attend} &I3D &ResNet-152 &Faster R-CNN &28.2M &\xmark 
&- &- &- &- &- &35.2 \\ 
% TSN\cite{yang2019question} &C3D &VGG16 &\xmark &- &\xmark 
% &27.9 &46.1 &84.1 &77.8 &37.6 &35.4 \\ 
% TSN 多步推理 
& HGA\cite{jiang2020reasoning} &C3D &VGG16 &\xmark &121.4M &\xmark 
&29.2 &45.7 &83.5 &75.2 &34.0 &35.5 \\
& DualVGR\cite{wang2021dualvgr} &ResNeXt-101 &ResNet-101 &\xmark &34.1M &\xmark 
&29.4 &45.6 &79.8 &76.7 &36.4 &35.5 \\ 

% conditional relation (通过对不同尺度/帧数的feature采样实现relation的表征)
&Park~\etal\cite{park2021bridge} &ResNeXt-101 &ResNet-101 &\xmark &- &\xmark 
&- &- &- &- &- &36.9 \\ % GNN for both interaction and fusion 
% &TPT\cite{peng2021temporal} &ResNeXt-101 &ResNet-101 &\xmark &- &\xmark 
% &- &- &- &- &- &37.7 \\ 
\rowcolor{gray!15}\cellcolor{white}
& \textbf{$\ModelName\text{-V}$} &\xmark &ResNet-101 &Faster R-CNN &10.7M &\xmark &34.1 &50.9 &81.5 &\textbf{82.8} &42.2 &40.6 \\ 
\cline{2-13}
\rowcolor{gray!15}\cellcolor{white} 
& \multicolumn{1}{l|}{\textbf{$\ModelName$ (ALL)}} &\xmark &ResNet-101 &Faster R-CNN &15.0M &\xmark &\textbf{50.3} &\textbf{77.1} &\textbf{94.2} &81.3 &\textbf{48.4} &\textbf{59.4} \\ 
\hline 
\end{tabular}
}
\end{center}
\end{table*}
 
% ***********************************************************************
\subsection{Experimental Settings} 

\subsubsection{Evaluation Datasets}
We evaluate the proposed $\ModelName$ framework on two VideoQA benchmarks, MSRVTT-QA~\cite{xu2017video}, and KnowIT VQA~\cite{garcia2020knowit}. 
MSRVTT-QA provides captions related to video content and KnowIT VQA provides subtitles (sub) and highly structure knowledge (know). 
These annotated captions, subtitles, and knowledge serve as the extra linguistic inputs and generate multi-grained linguistic representations. 
Table~\ref{tab:data_statistics} summarizes statistics of the experimental datasets. 
Specifically, MSRVTT-QA has 10K videos and 243,680 question-answer pairs. The question setting is open-ended, and the size of the pre-defined answer set is 1000. There are five questions types: What, Who, How, When, and Where. 
KnowIT VQA is a small-scale multiple-choice VideoQA dataset comprised of 12,087 video clips and 24,282 question-answer pairs. It provides four candidate answers for each question. There are four questions types: Visual (Vis.), Textual (Text.), Temporal (Temp.), and Knowledge (Know.).

\subsubsection{Details of Feature Extraction} 
\label{sec:data_preprocess} 
To obtain the inputs of $\ModelName$ (\ie, the visual and linguistic features), we utilize ResNet-101~\cite{he2016deep} pre-trained on ImageNet~\cite{deng2009imagenet} to extract the holistic image appearance features for all experimental datasets, and utilize bottom-up attention Faster R-CNN~\cite{anderson2018bottom} pre-trained on Visual Genome~\cite{krishna2017visual} to detect objects and corresponding class-attributes in each sampled image. 
More specifically, for MSRVTT-QA, we sampled 64 frames at an equal interval from each video clip, and 12 sentences as the linguistic sentences for each video clip from the provided caption annotations in~\cite{xu2016msr}. The number of detected objects in each sampled image is 10. 
For KnowIT VQA, the number of sampled images and detected objects are 32 and 12, respectively. We use original subtitles and the provided knowledge condensed from subtitles of video as the linguistic sentences.

\subsubsection{Implementation Details} 
For $\ModelName$, we set the standard feature dimensionality $d$ to 512, the layer $l$ of GCN to 1. The number of semantic role $N_r$ in Sec. \ref{subsec:SRG} is 16, and the number of remained maximum values $N_n$ for matrix $\Amat_{se}$ in Eq.~(\ref{eq:graph_learner}) is set to 5. 
The number of heads $N_h$ in Eq.~(\ref{eq:alpha_ij}) is respectively set to 16 and 8 on MSRVTT-QA and KnowIT-VQA. 
We implement $\ModelName$ on two NVIDIA GeForce GTX 2080Ti GPUs, and utilize AdamW optimizer~\cite{loshchilov2017decoupled} with an initial learning rate of 8e-5 and a batch size of 256 for 80 epochs. 
Code will be available at \url{https://github.com/jingjing12110/LiVLR-VideoQA}. 

\subsection{Comparisons with State-of-the-Arts} 
We compare $\ModelName$ with state-of-the-arts on an open-ended (MSRVTT-QA) and a multi-choice (KnowIT VQA) datasets.

\begin{table}[!t]
\begin{center}
\caption{Comparisons with state-of-the-arts on KnowIT VQA. 
G (H) indicates that the linguistic input knowledge is generated online (human-annotated offline). 
}
\label{tab:knowit}
\vspace{-2mm}
\setlength{\tabcolsep}{1.mm}{
\small 
\begin{tabular}{l|c|cc|cccccc}
\hline
\multirow{2}{*}{Method} 
&\multirow{2}{*}{\makecell[c]{Vis.\\Input}} 
&\multicolumn{2}{c|}{Ling. Input} 
&\multicolumn{5}{c}{Accuracy (\%)} 
\\ 
& &\footnotesize{Sub.} &\footnotesize{Know.} 
&\footnotesize{Vis.} &\footnotesize{Text.} &\footnotesize{Temp.} &\footnotesize{Know.} &\footnotesize{All} 
\\
\hline 
\hline 
% Human &- &96.1 &93.6 &85.7 &86.7 &89.6 \\ 
% \hline 
TVQA\cite{lei2018tvqa} &concept &\checkmark &\xmark 
&61.2 &64.5 &54.7 &46.6 &52.2 \\ 
ROCK\cite{garcia2020knowit} &image &\checkmark &G 
&65.4 &68.1 &62.8 &64.6 &65.2 
\\ 
ROCK\cite{garcia2020knowit} &concept &\checkmark &G 
&65.4 &68.5 &62.8 &64.6 &65.2 
\\
ROCK\cite{garcia2020knowit} &facial &\checkmark &G 
&65.4 &68.8 &62.8 &64.6 &65.2 
\\
ROCK\cite{garcia2020knowit} &caption &\checkmark &G 
&64.7 &67.8 &59.3 &64.3 &64.6 
\\
ROLL\cite{garcia2020knowledge} &des. &\checkmark &G 
&71.8 &73.9 &64.0 &71.3 &71.5 
\\
\hline 
ROLL~\cite{garcia2020knowledge} &des. &\checkmark &H 
&70.8 &75.4 &57.0 &56.7 &62.0 
\\
ROCK\cite{garcia2020knowit} &concept &\checkmark &H 
&74.7 &\textbf{81.9} &75.6 &70.8 &73.1 
\\
\rowcolor{gray!15} \textbf{$\ModelName$} &I + O &\checkmark &\xmark 
&76.4 &75.1 &72.2 &77.3 &77.0 
\\ 
\rowcolor{gray!15} \textbf{$\ModelName$} &I + O &\xmark &H 
&\textbf{79.3} &70.5 &\textbf{76.4} &\textbf{78.0} &\textbf{77.1} 
\\ 
\hline
\end{tabular}
}
\end{center}
\end{table}

% ***********************************************************************
\subsubsection{Comparisons on MSRVTT-QA} 
For MSRVTT-QA, we compare the proposed $\ModelName$ with recent methods, including Park~\etal\cite{park2021bridge}, DualVGR\cite{wang2021dualvgr}, HGA\cite{jiang2020reasoning}, MASN\cite{seo2021attend}, HCR\cite{le2020hierarchical}, MiNOR\cite{jin2019multi}, HME\cite{fan2019heterogeneous}, GRA\cite{xu2017video}, Co-mem\cite{gao2018motion}, ST-VQA\cite{jang2017tgif}, VQA-T\cite{yang2021just}, CoMVT\cite{seo2021look}, ClipBERT\cite{lei2021less}, and SSML\cite{amrani2021noise}. 
It is worth noting that VQA-T, CoMVT, ClipBERT, and SSML (\ding{172}) adopt large-scale video-language pretraining to enhance the downstream VideoQA task. 
Generally, the performance of pretraining-based methods is better than the performance of those methods without pretraining. 
ST-VQA, Co-mem, GRA, HME, MiNOR, and HCR (\ding{173}) use attention mechanism to achieve cross-modal representations interaction and fusion. 
MASN, HGA, DualVGR, and Park~\etal (\ding{174}) adopt graph neural networks (GNNs). 
Specifically, MASN adopts GNNs to encode visual representations. 
HGA sequentially applies the attention mechanism and GNNs for representations fusion. 
DualVGR and Park~\etal are most similar with the proposed $\ModelName$ utilizing GNNs for both representation encoding and fusion. 

Table~\ref{tab:MSRVTT} summarizes the comparisons with aforementioned methods on MSRVTT-QA. 
Since all compared methods listed in Table~\ref{tab:MSRVTT} do not utilize extra-linguistic information in addition to the given question, for a fairer comparison, we mainly compare our \textbf{$\ModelName\text{-V}$} (\ie, \#2 in Table~\ref{tab:abl_comps}), which only integrates the obtained multi-grained visual representations $(\Xmat_{V, g}, \Xmat_{V, l})$ by the proposed $\ModuleName$, with these methods. 
Integrating only multi-grained visual representations, the performance of our method ($\ModelName\text{-V}$) has surpassed the best in the same group (\textbf{40.6\%} vs. 36.9\%). 
After using the extra multi-grained linguistic representations, that is, injecting linguistic representations $(\Xmat_{L, g}, \Xmat_{L, l})$ into $\ModuleName$, the overall performance of $\ModelName$ is further improved by 18.8$\%$ in the group (\ding{173}). 
Moreover, compared with the best pretraining-based method, VQA-T, the performance of our $\ModelName\text{-V}$ is also comparable.

% *********************************************************************
\begin{table}[!tbp]
\begin{center}
\caption{Comparison with alternative Representation Integration (RI) methods on MSRVTT-QA and KnowIT-VQ. 
}
\label{tab:abl_RI} 
\vspace{-2mm}
\small
\setlength{\tabcolsep}{1.mm}{
\begin{tabular}{l|ccc|c}
\hline
Dataset &RI-Concat &RI-AT &RI-GCN &\textbf{$\ModuleName$ (Ours)} \\
\hline
\hline
MSRVTT-QA\cite{xu2017video} &46.73 &52.35 &56.16 &\textbf{59.44} \\
KnowIT-VQA\cite{garcia2020knowit} &66.07 &69.01 &73.79 &\textbf{77.10} \\
\hline
\end{tabular}
} % scale box
\end{center}
\end{table}
% *********************************************************************
 
% ***********************************************************************
\begin{figure}[!t] 
\begin{center} 
\includegraphics[width=1.0\linewidth]{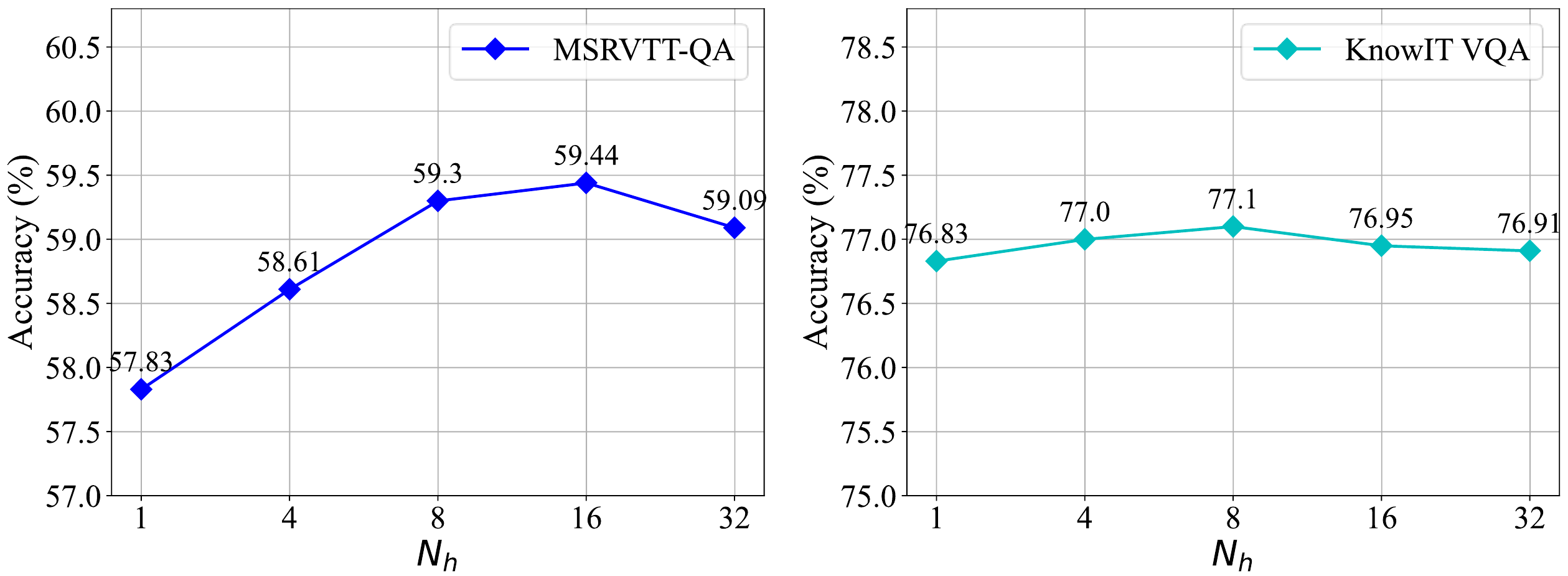}
\end{center}
\vspace{-4mm}
\caption{Ablation studies on the number of heads $N_h$ in Eq.(\ref{eq:alpha_ij}) on MSRVTT-QA~\cite{xu2017video} and KnowIT-VQA~\cite{garcia2020knowit}. 
}
\label{fig:vis_param}
\end{figure}

% **********************************************************************
\subsubsection{Comparisons on KnowIT VQA} 
For KnowIT VQA, we compare $\ModelName$ with the latest reported
results on KnowIT VQA (four different settings of ROCK~\cite{garcia2020knowit} and TVQA~\cite{lei2018tvqa}) and ROLL~\cite{garcia2020knowledge}. 
Specifically, ROCK adopts four different techniques to describe visual contents of video frames: 
(a) \textit{image}, image-level features extracted using ResNet50~\cite{he2016deep}. 
(b) \textit{concept}, bag-of-words representations of the objects and their attributes obtained using detector~\cite{anderson2018bottom}. 
(c) \textit{facial}, bag-of-faces representations of main characters in the clip detected with face detector~\cite{parkhi2015deep}. 
(d) \textit{caption}, representations of sentences describing the visual content of the frames and sentences are obtained using~\cite{xu2015show}. 
ROLL generates unsupervised video scene descriptions (\textit{des.}) as the visual input. 
Our $\ModelName$ utilizes image-level appearance features and object-level region features (\textit{I + O}) as the visual input. 
To obtain multi-grained linguistic representation, we respectively exploit the provided subtitles ($N_s$=12) and knowledge ($N_s$=1) as the original inputs of our Linguistic Encoder. 

Results are shown in Table~\ref{tab:knowit}. 
Overall, the proposed $\ModelName$ outperforms previous methods by a large margin. 
Particularly, $\ModelName$ improves the overall performance by approximately 4$\%$ with respect to the best performing ROCK. 
Comparing the two cases of $\ModelName$ (using subtitles/knowledge as the linguistic input), they achieve similar overall accuracy. 
However, using knowledge as linguistic input performs poorer than using subtitles in answering textual-based questions. 
The possible reason is that $\ModelName$ can only obtain one pair of multi-grained linguistic representations from the provided knowledge sentence for one image-question pair, whose number is far less than the number of obtained multi-grained visual representations causing the visual information to dominate the representation integration process and weaken the effect of linguistic information.

\subsection{Ablation Studies} 
\label{sec:ab_all}
We conduct ablation studies on MSRVTT-QA and KnowIT-VQA to demonstrate the effectiveness of key components in the proposed $\ModelName$.\footnote{In all ablation studies, we use knowledge of KnowIT-VQA to obtain multi-grained linguistic inputs for $\ModelName$.} 

% **********************************************************************
% *********************************************************************
\begin{table}[!t]
\begin{center}
\caption{Explanations of crucial notations used in ablation studies. 
}
\label{tab:abl_comps_explanation}
\vspace{-2mm}
\small
\setlength{\tabcolsep}{1.8mm}{
\begin{tabular}{c|l}
\hline
Notation &\multicolumn{1}{c}{Explanation} \\ 
\hline
\hline
$\Xmat_{V, g}$ &The holistic visual representation. \\
$\Xmat_{V, l}$ &The fine-grained visual representation. \\
$\Xmat_{L, g}$ &The holistic linguistic representation. \\
$\Xmat_{L, l}$ &The fine-grained linguistic representation. \\
% $\ev_{no}$ & Using vanilla GCN (\ie, RI-GCN) for RI.\\ 
$\ev_{no}$ &Without using diversify-aware embeddings in $\ModuleName$.\\  
$\ev_{da}$ &Using diversify-aware embeddings in $\ModuleName$. \\
% $\ev_{ss}$ &Using source-sensitive embeddings in $\ModuleName$. \\
\hline
\end{tabular}
} 
\end{center}
\end{table}

% *********************************************************************
\begin{table}[!tbp]
% \vspace{-2mm} 
\begin{center}
\caption{Ablation studies on key components of $\ModelName$ on MSRVTT-QA (\textbf{M}) and KnowIT-VQA (\textbf{K}). 
}
\label{tab:abl_comps} 
\vspace{-2mm}
\small
\newcommand\myslbox{\diagbox[dir=SW, width=16em,height=\line]{\phantom{x}}{\phantom{x}}}
\setlength{\tabcolsep}{1.2mm}{
\begin{tabular}{c|c|cc|cc|cc|ccc}
\hline
\multicolumn{2}{c|}{\multirow{2}{*}{S/N}} 
&\multicolumn{2}{c|}{VEnc.}
&\multicolumn{2}{c|}{LEnc.} 
&\multicolumn{2}{c|}{$\ModuleName$} 
&\multirow{2}{*}{M\cite{xu2017video}} 
&\multirow{2}{*}{K\cite{garcia2020knowit}} 
% &\textbf{~M} &\textbf{K}
\\
% \cline{3-8}
\multicolumn{2}{c|}{~} 
&\scriptsize{$\Xmat_{V, g}$} &\scriptsize{$\Xmat_{V, l}$} 
&\scriptsize{$\Xmat_{L, g}$} &\scriptsize{$\Xmat_{L, l}$} 
&$\ev_{no}$ &$\ev_{da}$ 
& & \\ 
\hline
\hline
\multicolumn{2}{c|}{Ques-only}
&\multicolumn{6}{c|}{------} &31.20 &50.12 \\
\hline 
\parbox[t]{2mm}{\multirow{4}{*}{\rotatebox[origin=c]{90}{I}}} 
&\#1 &\checkmark &\checkmark & & 
&\checkmark & &38.99 &67.77 \\ 
&\#2 &\checkmark &\checkmark & & 
& &\checkmark &40.63 &70.21 \\
% \cdashline{1-9}
&\#3 & & &\checkmark &\checkmark 
&\checkmark & &49.45 &68.00 \\ 
&\#4 & & &\checkmark &\checkmark 
& &\checkmark &51.26 &68.73 \\
\hline
\parbox[t]{2mm}{\multirow{4}{*}{\rotatebox[origin=c]{90}{II}}} 
&\#5 &\checkmark & &\checkmark & 
&\checkmark & &48.12 &67.19 \\ 
&\#6 &\checkmark & &\checkmark & 
& &\checkmark &49.58 &70.00 \\
% \cdashline{1-9}
&\#7 & &\checkmark & &\checkmark 
&\checkmark & &50.03 &67.87 \\ 
&\#8 & &\checkmark & &\checkmark 
& &\checkmark &51.85 &71.98 \\
\hline
\parbox[t]{2mm}{\multirow{2}{*}{\rotatebox[origin=c]{90}{III}}} 
&\#9 &\checkmark &\checkmark &\checkmark &\checkmark 
&\checkmark & &56.16 &73.79 \\ 
&\#10 &\checkmark &\checkmark &\checkmark &\checkmark 
& &\checkmark &\textbf{59.44} &\textbf{77.10} \\
\hline
\end{tabular}
} % tab col sep
\end{center}
\end{table}

% **********************************************************************
\begin{table}[!tbp]
\begin{center}
\caption{Comparisons of respectively using GCN and FC as fine-grained visual and linguistic representation extractor on MSRVTT-QA (\textbf{M}) and KnowIT-VQ (\textbf{K}). \#9 and \#10 are the same with that in Table~\ref{tab:abl_comps}.
}
\label{tab:abl_GCN2FC} 
\vspace{-2mm}
\small
\setlength{\tabcolsep}{2.0mm}{
\begin{tabular}{l|cc|cc|cc|cc}
\hline
\multirow{2}{*}{S/N} 
&\multicolumn{2}{c|}{VEnc.}
&\multicolumn{2}{c|}{LEnc.} 
&\multicolumn{2}{c|}{$\ModuleName$} 
&\multirow{2}{*}{\textbf{M}\cite{xu2017video}} 
&\multirow{2}{*}{\textbf{K}\cite{garcia2020knowit}} 
\\ 
&\footnotesize{GCN} &\footnotesize{FC} &\footnotesize{GCN} &\footnotesize{FC} &$\ev_{no}$ &$\ev_{da}$ & & \\
\hline
\hline
\#11 & &\checkmark & &\checkmark &\checkmark & &40.32 &68.12 \\ 
\#9 &\checkmark & &\checkmark & &\checkmark & &56.16 &73.79 \\
\#12 & &\checkmark & &\checkmark & &\checkmark &52.96 &73.01 \\
\#10 &\checkmark & &\checkmark & & &\checkmark &59.44 &77.10 \\
\hline 
\end{tabular}
} % scale box
\end{center}
\end{table}

% *********************************************************************

\subsubsection{Effectiveness of the proposed RI Method ($\ModuleName$)} 
In our $\ModelName$, $\ModuleName$ is designed to better integrate multi-grained visual and linguistic representations. 
To evaluate its effectiveness, we compare $\ModuleName$ with three alternative methods of representation integration (RI) on the above two benchmarks. 
Specifically, $\blacktriangleright$RI-GCN: using a vanilla GCN to integrate the obtained question-related multi-grained visual and linguistic representations \{$\Xmat_{V, g}, \Xmat_{V, l}, \Xmat_{L, g}, \Xmat_{L, l}$\}$^q$. 
% RI-GCN is the existing best RI method similar to our proposed $\ModuleName$ and is originally proposed in~\cite{jiang2020reasoning} to integrate cross-modality representations. 
RI-GCN is the most similar method to our proposed $\ModuleName$. 
However, RI-GCN does not encode the diversity-aware information for graph $\mathcal{G}_{da}$. 
$\blacktriangleright$RI-AT: integrating \{$\Xmat_{V, g}, \Xmat_{V, l}, \Xmat_{L, g}, \Xmat_{L, l}$\}$^q$ using the co-attention operation like the work in HGA~\cite{jiang2020reasoning}. 
$\blacktriangleright$RI-Concat: integrating the obtained representations \{$\Xmat_{V, g}, \Xmat_{V, l}, \Xmat_{L, g}, \Xmat_{L, l}$\}$^q$ by a vector concatenation operation. 

Results are shown in Table~\ref{tab:abl_RI}. 
The large performance gap with respect to the three alternative RI methods suggests the effectiveness of our proposed $\ModuleName$. 
Furthermore, compared with the best alternative RI method (RI-GCN), our proposed $\ModuleName$ can improve the performance by 3.28$\%$ (59.44 vs. 56.16) and 3.31$\%$ (77.10 vs. 73.79) on MSRVTT-QA and KnowIT-VQA, respectively. 
The performance gains on two datasets demonstrate that encoding diversity-aware information is significant for the integration of multi-grained visual and linguistic representations.

\begin{figure*}[!t]
\begin{center}
% \subfigtopskip=2pt %设置子图与上面正文或别的内容的距离
\subfigbottomskip=1pt %设置第二行子图与第一行子图的距离，即下面的头与上面的脚的距离
\subfigcapskip=-5pt %设置子图与子标题之间的距离
\subfigure[Answering the questions based on the holistic and fine-grained visual content corresponding to the same video clip.]{
\includegraphics[width=18cm]{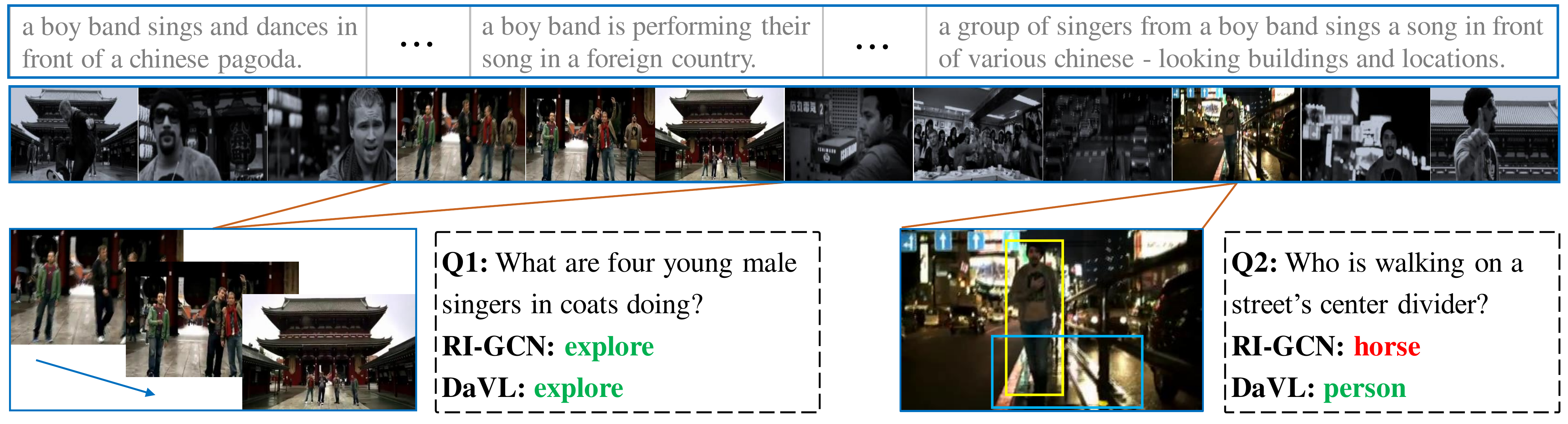}
\label{a1}
}
\subfigure[Answering the questions based on the visual and linguistic content corresponding to the same video clip.]{
\includegraphics[width=18cm]{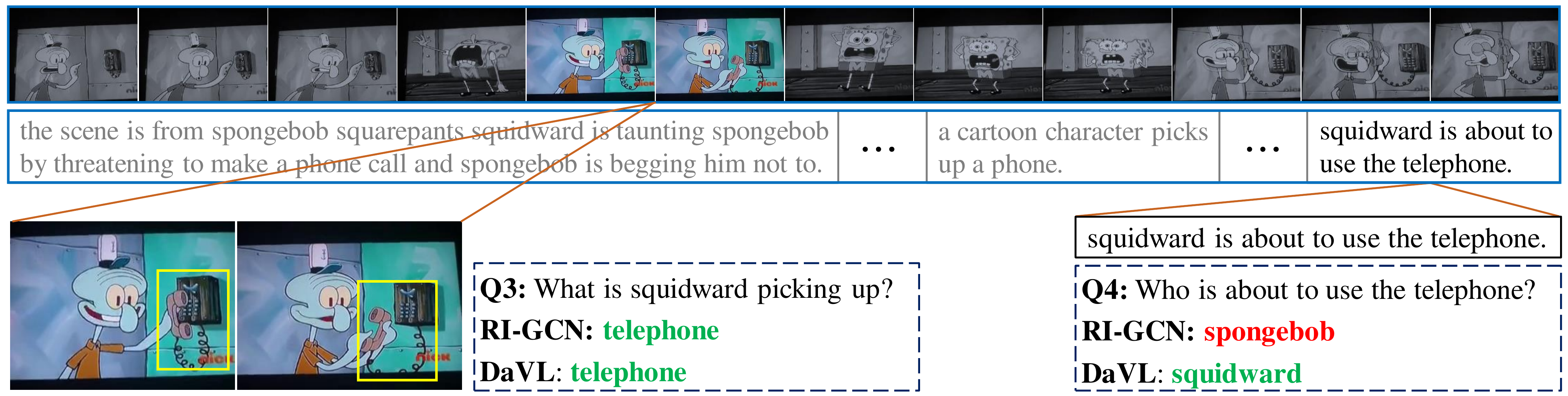}
\label{a2}
} 
\caption{Visualization examples of the comparison between two representation integration (RI) method on MSRVTT-QA~\cite{xu2017video}. 
% RI-GCN and our $\ModelName$
$\ModuleName$ is our proposed RI method and RI-GCN is the previous graph-based RI method. 
The \textcolor{red}{wrong} and \textcolor{ForestGreen}{correct} answers are highlighted in \textcolor{red}{red} and \textcolor{ForestGreen}{green}, respectively. 
}
\label{fig:vis_correct}
\end{center}
\end{figure*}
 
% **********************************************************************
\subsubsection{Effectiveness of $\ModuleName$ for Multi-grained Representations} 
To demonstrate that our proposed $\ModuleName$ is also effective for integrating multi-grained representations derived from a single source, we conduct the comparisons in Table~\ref{tab:abl_comps} (I). 
$\blacktriangleright$ \#1 vs. \#2: using the RI-GCN and our $\ModuleName$ respectively to integrate multi-grained visual representations ($\Xmat_{V,g}, \Xmat_{V, l}$). 
$\blacktriangleright$ \#3 vs. \#4: using the RI-GCN and our $\ModuleName$ respectively to integrate multi-grained linguistic representations ($\Xmat_{L, g}, \Xmat_{L, l}$). 
Results in Table~\ref{tab:abl_comps} (I) suggest that $\ModuleName$ is also effective for integrating single-source multi-grained representations.

% **********************************************************************
\subsubsection{Effectiveness of $\ModuleName$ for Cross-modal Representations} 
To evaluate the effectiveness of our $\ModuleName$ in integrating single-granularity cross-modal representations, we conduct the comparisons in Table~\ref{tab:abl_comps} (II). 
$\blacktriangleright$ \#5 vs. \#6: using the RI-GCN and our $\ModuleName$ respectively to integrate holistic cross-modal representations ($\Xmat_{V, g}, \Xmat_{L, g}$). 
$\blacktriangleright$ \#7 vs. \#8: using the RI-GCN and our $\ModuleName$ respectively to integrate fine-grained visual and linguistic representations ($\Xmat_{V, l}, \Xmat_{L, l}$). 
Results in Table~\ref{tab:abl_comps} (II) illustrate that the proposed $\ModuleName$ is also effective for single-granularity cross-modal representation integration.

% **********************************************************************
\subsubsection{Impact of Multi-grained Visual and Linguistic Representations} 
% The proposed $\ModelName$ values both visual and linguistic content at the same level by yielding four representations $\Xmat_{V, g}, \Xmat_{V, l}, \Xmat_{L, g}, \Xmat_{L, l}$. 
The proposed $\ModelName$ encodes visual and linguistic content by Visual and Linguistic Encoders with similar architectures. 
This can guarantee to some extent that the obtained holistic (fine-grained) representations from different modalities are at the same semantic level (sentence $\leftrightarrow$ image, semantic roles $\leftrightarrow$ object instances). 
To analyze the impact of multi-grained visual and linguistic representations, we first consider the two comparisons: 
$\blacktriangleright$ Table~\ref{tab:abl_comps} \#9 vs. \#1 vs. \#3 and $\blacktriangleright$ Table~\ref{tab:abl_comps} \#10 vs. \#2 vs. \#4.  
From the results in the table, we can observe that the performance of considering multi-grained visual and linguistic representations (\#9/\#10) is markedly better than the performance of using the single visual (\#1/\#2) or linguistic (\#3/\#4) representations. 
Secondly, we consider the two comparisons: 
$\blacktriangleright$ Table~\ref{tab:abl_comps} \#9 vs. \#5 vs. \#7 and $\blacktriangleright$ Table~\ref{tab:abl_comps} \#10 vs. \#6 vs. \#8. 
Analogously, we find that the performance of considering multi-grained visual and linguistic representations (\#9/\#10) is markedly better than the performance of using the single holistic (\#5/\#6) or fine-grained (\#7/\#8) representations. 
% Surprisingly, without our proposed $\ModuleName$, only encoding visual and linguistic content at the same semantic level, we have achieved state-of-the-art performance on the two datasets, indicating that the proposed $\ModelName$ establishes a strong baseline for the $\Problem$ task. 
Finally, to further illustrate the superiority of multi-grained visual and linguistic representations, especially the fine-grained visual and linguistic representations, we conduct the following experiment: $\blacktriangleright$ replacing the GCN in Visual and Linguistic Encoders with a two-layer FC network to obtain fine-grained visual representations. 
The results on MSRVTT-QA and KnowIT-VQA are shown in Table~\ref{tab:abl_GCN2FC}, which suggests that obtaining fine-grained visual and linguistic representations that encode relationships between visual objects or linguistic components is crucial.

\subsubsection{Hyperparameter} 
To conduct more detailed parameter analysis, we consider the key hyperparameter $N_h$ in Eq.~(\refeq{eq:alpha_ij}), which may directly affect the effectiveness of the proposed RI method $\ModuleName$. 
Specifically, the question-related attention block ($\mathcal{Q}_{\text{att}}$) is employed to associate the multi-grained visual and linguistic representations ($\Xmat_{V, g}, \Xmat_{V, l}, \Xmat_{L, g}, \Xmat_{L, l}$) to the question-related representation ($\Qmat$). This is significant to distill the question-related information from multi-grained visual and linguistic information. 
Specifically, we consider following settings: $N_h = 1, 4, 8, 16, 32$. 
From the experimental results in Figure~\ref{fig:vis_param}, we observe that, compared with the overall performance improvement, the performance fluctuation of $\ModelName$ using different $N_h$ is slight, demonstrating that our method is robust to the hyperparameter $N_h$. 

% **********************************************************************
\begin{figure}[!tbp]
\begin{center}
\includegraphics[width=1.0\linewidth]{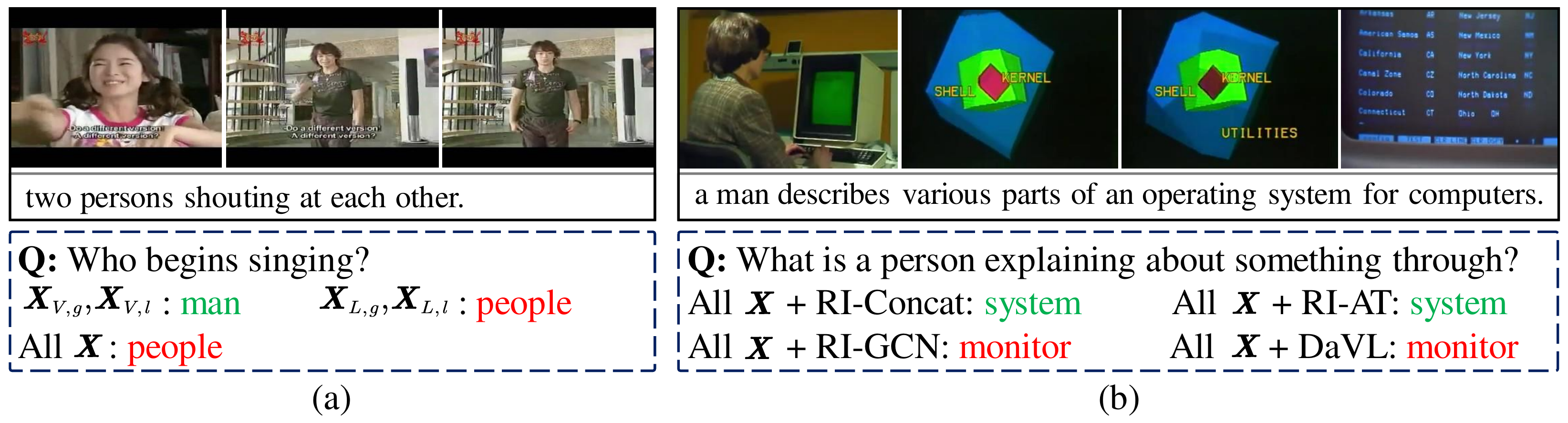}
\end{center}
\vspace{-2mm}
\caption{Two failure cases on MSRVTT-QA~\cite{xu2017video}. 
(a) A failure case caused by multi-grained linguistic representations in $\ModuleName$. 
(b) A failure case caused by graph-based RI methods. 
The \textcolor{red}{wrong} and \textcolor{ForestGreen}{correct} answers are respectively highlighted in \textcolor{red}{red} and \textcolor{ForestGreen}{green}. 
} 
\label{fig:vis_error}
\end{figure}
% ********************************************************************** 

\subsection{Qualitative Results} 

\subsubsection{Qualitative Examples} 
To qualitatively evaluate the effectiveness of the proposed representation integration method ($\ModuleName$), we visualize some prediction examples on MSRVTT-QA~\cite{xu2017video} in Figure~\ref{fig:vis_correct}. 
Specifically, in Figure~\ref{fig:vis_correct} (a), we show two questions corresponding to the same video stream. 
Answering Q1 needs to understand the holistic visual content described in a video clip. 
Answering Q2 needs to understand the fine-grained visual content described in one frame of the video. 
Both using RI-GCN and the proposed $\ModuleName$ can correctly answer the Q1, but using RI-GCN answers Q2 incorrectly. 
In Figure~\ref{fig:vis_correct} (b), although both using RI-GCN and $\ModuleName$ correctly answer the Q3 related to the visual content, using RI-GCN can not answer the Q4 related to the linguistic content. 
The two groups of comparisons between RI-GCN and $\ModuleName$ can demonstrate the effectiveness of the learnable index embeddings in graph-based representation integration, and the embeddings, to some extent, adaptively choose the needed representations for the specific question.

% \subsubsection{Visualization of Representation Correlation} 
% \begin{figure*}[!thp] 
% \begin{center} 
% \includegraphics[width=0.98\linewidth]{figure/correlation.pdf}
% \end{center}
% \vspace{-2mm}
% \caption{Visualization of representation correlation. (a) self-correlation of node representations before $\ModuleName$. (b) self-correlation of node representations after $\ModuleName$. (c) cross-correlation between node representations before $\ModuleName$ and node representations after $\ModuleName$. 
% }
% \label{fig:vis_correlation}
% \end{figure*}
% The quantitative results in Sec.\ref{sec:ab_all} validate the effectiveness of the proposed representation integration method $\ModuleName$. 
% But how does $\ModuleName$ work? 
% To analyze the possible explanation, we show the correlation (\ie, the cosine similarity) between node representations of the graph $\mathcal{G}_{gs}$ before/after encoding by $\ModuleName$. 
% % The results are shown in Figure~\ref{fig:vis_correlation}. 
% The comparison between Figure~\ref{fig:vis_correlation} (a) and (b), and the  cross-correlation map Figure~\ref{fig:vis_correlation} (c) illustrate how $\ModuleName$ integrates the obtained multi-grained visual and linguistic representations. 
% In Figure~\ref{fig:vis_correlation} (a), we hardly observe correlation among $\Xmat_{V, g}, \Xmat_{V, l}, \Xmat_{L, g}, \Xmat_{L, l}$. 
% With GSNet, Figure~\ref{fig:vis_correlation} (b), the representation correlation is learned and discriminability is also preserved to some extent. 

\subsubsection{Failure Cases Analysis} 
Figure~\ref{fig:vis_error} shows two failure cases on MSRVTT-QA. 
In Figure~\ref{fig:vis_error} (a), considering the multi-grained linguistic representations in $\ModuleName$, our proposed $\ModelName$ instead answers the question incorrectly. 
The case suggests that the learnable index embeddings may not be enough for $\ModuleName$ to select the needed visual representations and ignore the irrelevant linguistic representations when answering the question only related to the visual content. 
In Figure~\ref{fig:vis_error} (b), using graph-based RI methods (\ie, RI-GCN and $\ModuleName$) to integrate multi-grained visual and linguistic representations, the VideoQA model answers the question incorrectly. 
While the model answers the question correctly when using the other two simple RI methods. 
The case shows that graph-based RI methods sometimes may lose the discriminability between nodes (\ie, different types of representations) when answering the semantic-complicated question that needs to jointly understand visual and linguistic content, which is the inherent trouble that graph-based RI methods will cause. 

% the irrelevant linguistic representations sometimes may mislead $\ModuleName$ and the discriminability of the learnable index embeddings 

\section{Conclusion}
In this paper, we propose a Lightweight Visual-Linguistic Reasoning framework ($\ModelName$), which mainly consists of Visual Encoder, Linguistic Encoder, and the devised Diversity-aware Visual-Linguistic Reasoning module ($\ModuleName$). 
Specifically, $\ModelName$ first adopts the Visual and Linguistic Encoders to obtain multi-grained visual and linguistic representations, and then utilizes $\ModuleName$ to integrate the obtained representations and yield a joint representation for answer prediction. 
Extensive ablation studies are conducted to explore the performance contribution of the crucial components of $\ModelName$. 
The proposed $\ModelName$ is lightweight and shows its superiority on an open-ended and a multiple-choice VideoQA datasets. 
In the future, we aim to explore a new representation integration method that can be more flexible in selecting the needed representations according to the given question.

\bibliographystyle{IEEEtran}
\bibliography{Reference}

\end{document}